\documentclass[letterpaper, 10 pt, conference]{ieeeconf}  

\IEEEoverridecommandlockouts                              

\title{\LARGE \bf
Region-Based Planning for 3D Within-Hand-Manipulation via Variable Friction Robot Fingers and Extrinsic Contacts
}

\author{Alp Sahin$^{1}$, Adam J. Spiers$^{2}$ and Berk Calli$^{1}$
\thanks{$^{1}$Authors are with Robotics Engineering Program, Worcester Polytechnic Institute, 85 Prescott Street, Worcester, MA-01605, USA
        {\tt\small \{asahin;bcalli\}@wpi.edu}}%
\thanks{$^{2}$Author is with the Electrical and Electronic Engineering Department, Imperial College London, London, SW7 2AZ, United Kingdom
        {\tt\small a.spiers@imperial.ac.uk}}%
\thanks{This work was supported in part by the National Science Foundation under grant IIS-1900953.}%
}

\usepackage[vlined,ruled]{algorithm2e}
\SetKwProg{Fn}{Function}{}{}
\usepackage{cite}
\usepackage{amsmath,amssymb,amsfonts}
\usepackage{algorithmic}
\usepackage{graphicx}
\usepackage{textcomp}
\usepackage{xcolor}
\usepackage{gensymb}
\usepackage[utf8]{inputenc}
\usepackage{array}
\usepackage{wrapfig}
\usepackage{multirow}
\usepackage{tabu}
\usepackage{fullpage}
\usepackage{times}
\let\labelindent\relax
\usepackage{enumitem}
\setlist[itemize]{leftmargin=*}
\setlist[enumerate]{leftmargin=*}
\usepackage{fancyhdr,graphicx,amsmath,amssymb}
\usepackage[ruled,vlined]{algorithm2e}
\include{pythonlisting}
\usepackage{lipsum}
\usepackage{multicol}
\usepackage{todonotes}
\usepackage{booktabs}

\begin{document}

\maketitle
\thispagestyle{empty}
\pagestyle{empty}

\begin{abstract}

Attempts to achieve robotic Within-Hand-Manipulation (WIHM) generally utilize either high-DOF robotic hands with elaborate sensing apparatus or multi-arm robotic systems. In prior work we presented a simple robot hand with variable friction robot fingers, which allow a low-complexity approach to within-hand object translation and rotation, though this manipulation was limited to planar actions. In this work we extend the capabilities of this system to 3D manipulation with a novel region-based WIHM planning algorithm and utilizing extrinsic contacts. The ability to modulate finger friction enhances extrinsic dexterity for three-dimensional WIHM, and allows us to operate in the quasi-static level. The region-based planner automatically generates 3D manipulation sequences with a modified A* formulation that navigates the contact regions between the fingers and the object surface to reach desired regions. Central to this method is a set of object-motion primitives (i.e. within-hand sliding, rotation and pivoting), which can easily be achieved via changing contact friction. A wide range of goal regions can be achieved via this approach, which is demonstrated via real robot experiments following a standardized in-hand manipulation benchmarking protocol.

\end{abstract}

\section{INTRODUCTION}


Within-hand-manipulation (WIHM), which is also known as in-hand-manipulation (IHM), may be defined as changing the grasp on an object without the need for regrasping \cite{cruciani-ral2020-benchmarking-in-hand}. While in-grasp object pose modification  is a commonplace task for humans (e.g. when moving a grasped pen into a writing position), WIHM remains a challenge for robotic systems. Several approaches have been presented to address this problem, a number of which rely on the redundancy of high-DOF robotic fingers and extensive object state information, often obtained by elaborate sensing systems \cite{Dyn_Sliding_Manip, andrychowicz2020learning}. Such hardware requirements are often associated with high financial cost, low mechanical robustness and limited portability to real-world environments. 

Alternative approaches to WIHM using relatively simple grippers and sensors have been proposed: the method in \cite{Dex_Manip_Graph} adopts a dual arm robotic systems with 1 DOF parallel robot grippers; methods in \cite{Two_Phase_Gripper,Prehensile_Pushing_IHM} utilizes extrinsic dexterity strategies; methods in \cite{b25} leverages underactuation and visual feedback. Our approach to achieve robust WIHM relies on the ability of altering the surface friction of robot fingers \cite{VFF_original}. By changing the effective friction between the object and the finger, we were able to achieve controlled sliding and rotation of the object within-hand. Various derivations of this design has also been proposed \cite{origami2020,Gripper_Cloth_Manip}. However, these systems have so far only achieved these manipulations in a 2D plane.

\begin{figure}
    \includegraphics[width=8cm,height=7cm]{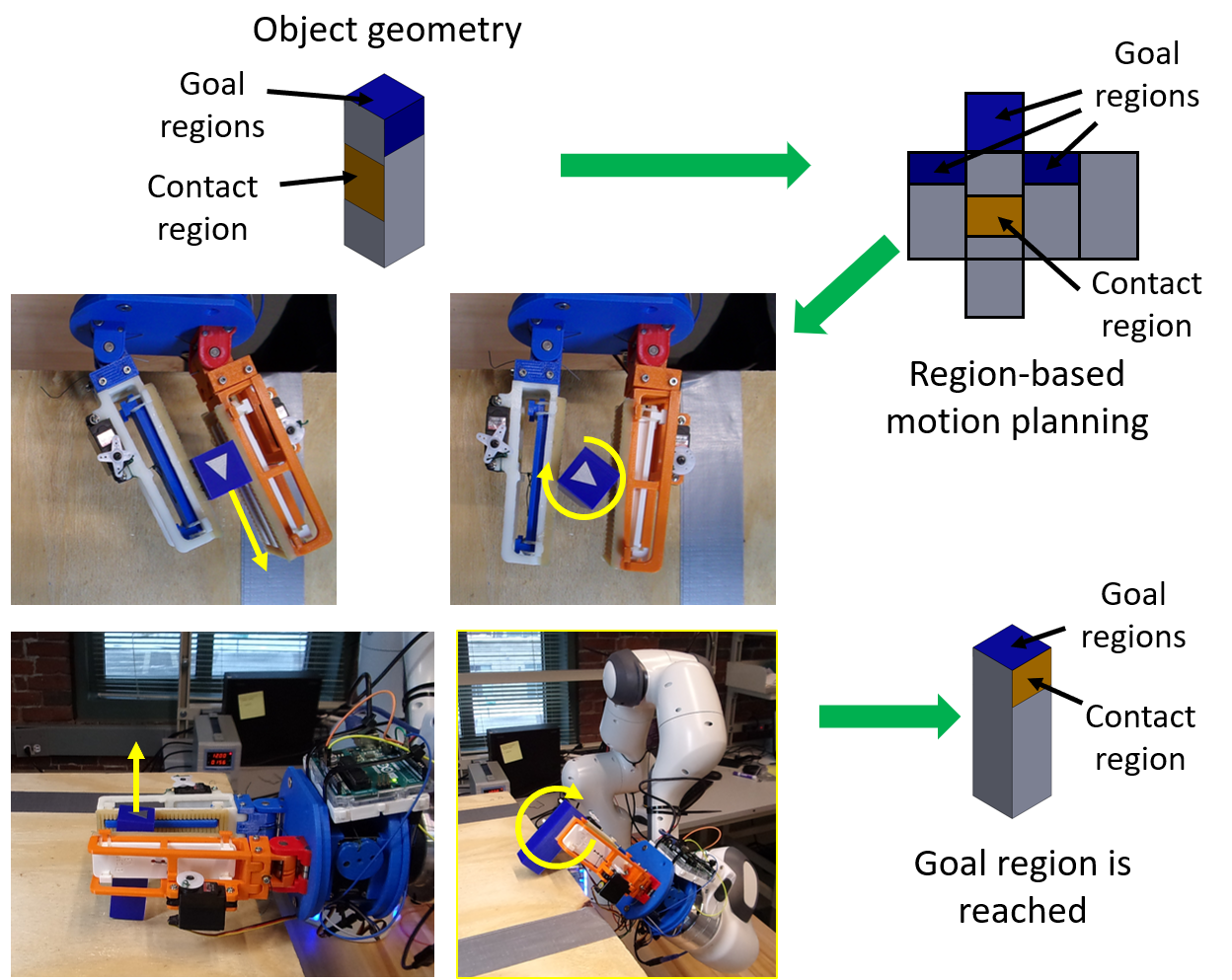}
    \vspace{-20pt}
    \caption{Proposed region-based motion planning algorithm generates a sequence of manipulation primitives, given the object geometry and goal regions. 3D manipulation setup makes use of controlled sliding and rotation via variable friction fingers and extrinsic dexterity based manipulation strategies.}
    \label{fig:WPI_Panda}
    \vspace{-20pt}
\end{figure}


This work's contributions are as follows (summarized in Fig. 1):

\textbf{A novel region-based WIHM method:} Different from the existing algorithms in literature, which formulate WIHM with point contacts, our planner works with contact areas and desired contact regions. We believe that this is not only a more realistic implementation for soft and/or flat finger surfaces (which are very common), but also a useful formulation for achieving various robotic tasks. For example, as humans when we grasp a key and re-position it within our hand, we do not have desired point locations that we want to achieve on the key. Rather, we shift our fingers to a particular region on the key, while leaving the tip region free. The proposed planner utilizes a very similar formulation.

\textbf{Quasi-static extrinsic dexterity:} We leverage the ability of being able to alter the friction at the contact locations for achieving extrinsic dexterity-based actions. Lowering the friction at the finger surfaces allows the robot to reliably pivot and slide the object within hand using extrinsic contacts via kinematics-level formulations, without the need of any force sensing or control.

By combining these two strategies, we are able to conduct automatic WIHM in 3D, i.e. move the fingers from an initial contact region to a given desired region. To the best of our knowledge, the proposed system surpasses the automatic WIHM abilities of any other system in the literature: the planner can utilize controlled sliding, rotation and pivoting action and achieve trajectories with significantly large region displacements. We demonstrate these abilities using a Franka Emika robot in Section~\ref{sec:experimental_results}. In this work, we assume that 3D geometry of the objects is known and the generated plans are executed in an open-loop manner. We discuss our future work regarding closed-loop implementations and possible mechanical improvements in Section~\ref{sec:discussion}.

\section{Related Work and Background}
\label{sec: litreview}
We first present the state of the art of WIHM systems. Following that, we summarize our prior work with variable friction hands and how it relates to this paper.

\subsection{WIHM Literature}

Dual arm robotic systems have been shown to achieve complex WIHM with only 1 DOF parallel robot grippers with gripping force regulation and a graph based planning approach \cite{Dex_Manip_Graph}. Given that few robot systems have the benefit of a second arm, in \cite{Prehensile_Pushing_IHM} a single robot arm with a modified parallel gripper pushed a grasped object against external fixtures to achieve WIHM, in a method known as \textit{extrinsic dexterity}. That paper relied on dynamic models of the object/gripper interaction to enable controlled slipping and pivoting of objects against different fingertip morphologies. Though many methods for WIHM are based on relatively complex dynamic models, more recently it has been shown that WIHM can be achieved with high-dexterity hands using only kinematic planning, which negates the need to have a-priori knowledge of the dynamic properties of objects \cite{sundaralingam2019relaxed}, thus permitting use in unstructured environments. However, the cost and fragility of the high-dexterity hand is high compared to simpler grippers, and can limit the adoption of such hardware into practical systems. Our approach has been to combine the benefits of extrinsic dexterity with kinematic planning via the combination of a robot arm and Variable Friction robot hand.

\subsection{Background for the Variable Friction Hand}

While the motion of human fingers is a major contributor to WIHM, another less-obvious feature is the material properties of the fingerpads themselves. Indeed, the mechanics of these elements are complex, resulting from ridges, sweat production and layers of different tissue types. These combined elements lead to a non-linear relationship between contact force and fingerpad surface friction \cite{tomlinson2007review}. The outcome is that humans have the ability to selectively grasp and slide objects against their fingers via the modification of contact force.

In our past work, we proposed a simple mechanism that enabled the object-contacting surface of a robot finger to be alternated between two interlaced elements with different coefficients of frictions, providing either gripping or sliding object interactions. This mechanism is presented in detail in \cite{VFF_original}. When implemented into a 2-finger, 2-DOF robot hand, it was demonstrated that despite a lack of kinematic redundancy, friction-modification enabled objects to be translated and rotated within the grasp of the hand. However, this object motion, which is based on the simultaneously motion of both 1-DOF fingers, is limited to one plane. 

In this work we make use of both the Variable Friction Hand's initial WIHM capabilities, while also exploiting it's ability to reduce friction, as a method of allowing the held object to pivot and slide when exposed to the external forces that result from extrinsic dexterity manipulations.

\section{3D Within-Hand Manipulation}
\label{sec:3d_wihm}

\begin{figure}[t!]
    \centering
    \includegraphics[width=\linewidth]{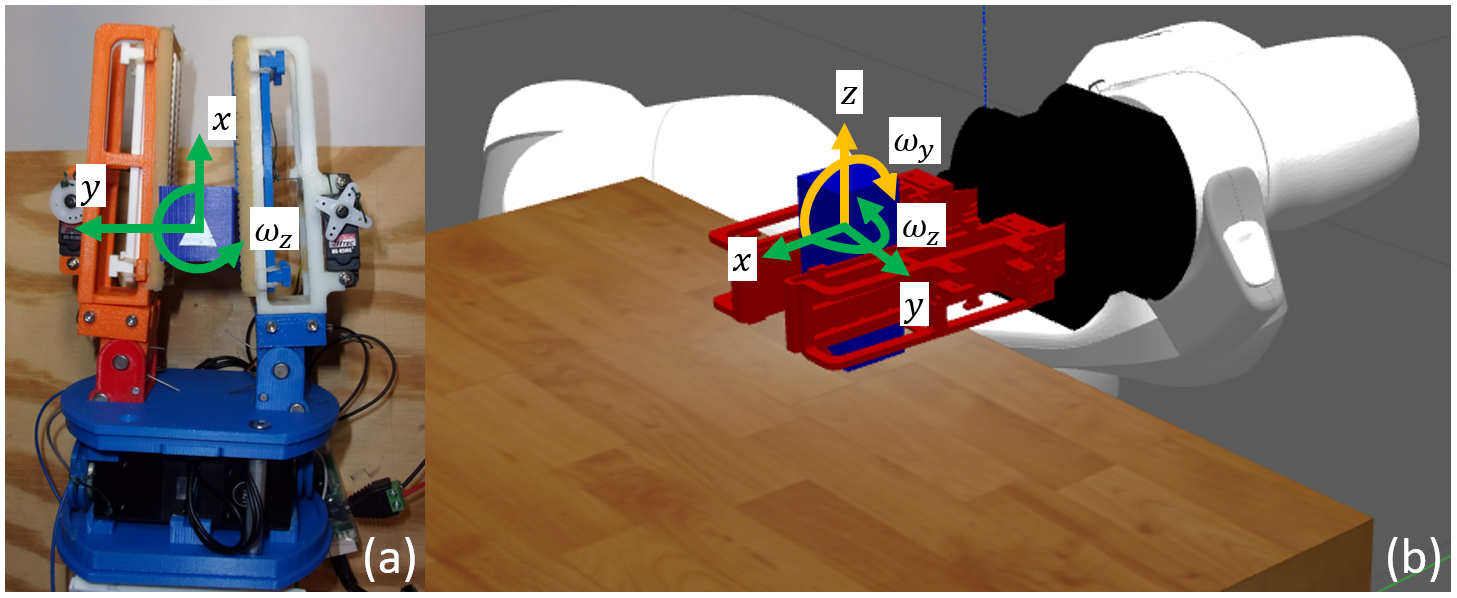}
    \vspace{-20pt}
    \caption{(a) In prior work, a 2-DOF robot hand was proposed that is capable of translation and rotation in 2D. (b) By combining this hand with a robotic arm, extrinsic dexterity based manipulation primitives such as prehensile pushing and pivoting can be utilized to achieve 3D manipulation.}
    \label{fig:DOF}
     \vspace{-20pt}
\end{figure}

In this work, we consider a within-hand manipulation task that requires the contact between the hand and the object to be navigated from an initial region to a goal region on the object. As shown in Fig. \ref{fig:DOF}a, previously used setup for the variable friction hand was restricted to 2D within-hand manipulation, where the object was able to translate and rotate within-hand along a plane (represented by green arrows). Using these manipulation primitives, only a small set of goal regions were reachable. Here, we expand the workspace to 3D to include the degrees of freedom shown in Fig. \ref{fig:DOF}b by extrinsic dexterity-based manipulation strategies (represented by yellow arrows). With the improved action space, we can reach a larger set of goal regions. Our region-based planner explained in Section~\ref{sec:region-based_planning} solves for a sequence of manipulation primitives given a manipulation task according to the provided problem formulation.

Following manipulation primitives are performed using extrinsic dexterity-based manipulation strategies, such as prehensile pushing, pivoting and exploitation of gravity, while assuming a horizontal support surface is available in the environment.

\subsection{Moving Contact Up and Down}
\label{sec:move_contact_up_down}

Contact between the fingers and the object is translated in positive $z$-direction (Fig.~\ref{fig:DOF}b) by exploiting the gravitational forces acting on the object. To achieve this action, both finger frictions are switched to the low friction state and fingers slide up on the object while the contact is maintained with the support surface. Thanks to the friction modulation, this can be achieved via a simple kinematic model, without using any force sensing or control. 

Likewise, having the contact friction low, contact between the fingers and the object is translated in negative $z$-direction through prehensile pushes against the supporting surface. Same contact assumptions and kinematic model hold for controlling the amount of pushing. After these operations the friction states are switched to high to recover the firm grip.

\subsection{Pivoting}
\label{sec:pivoting}

\begin{figure}[t!]
    \centering
    \includegraphics[width=\linewidth]{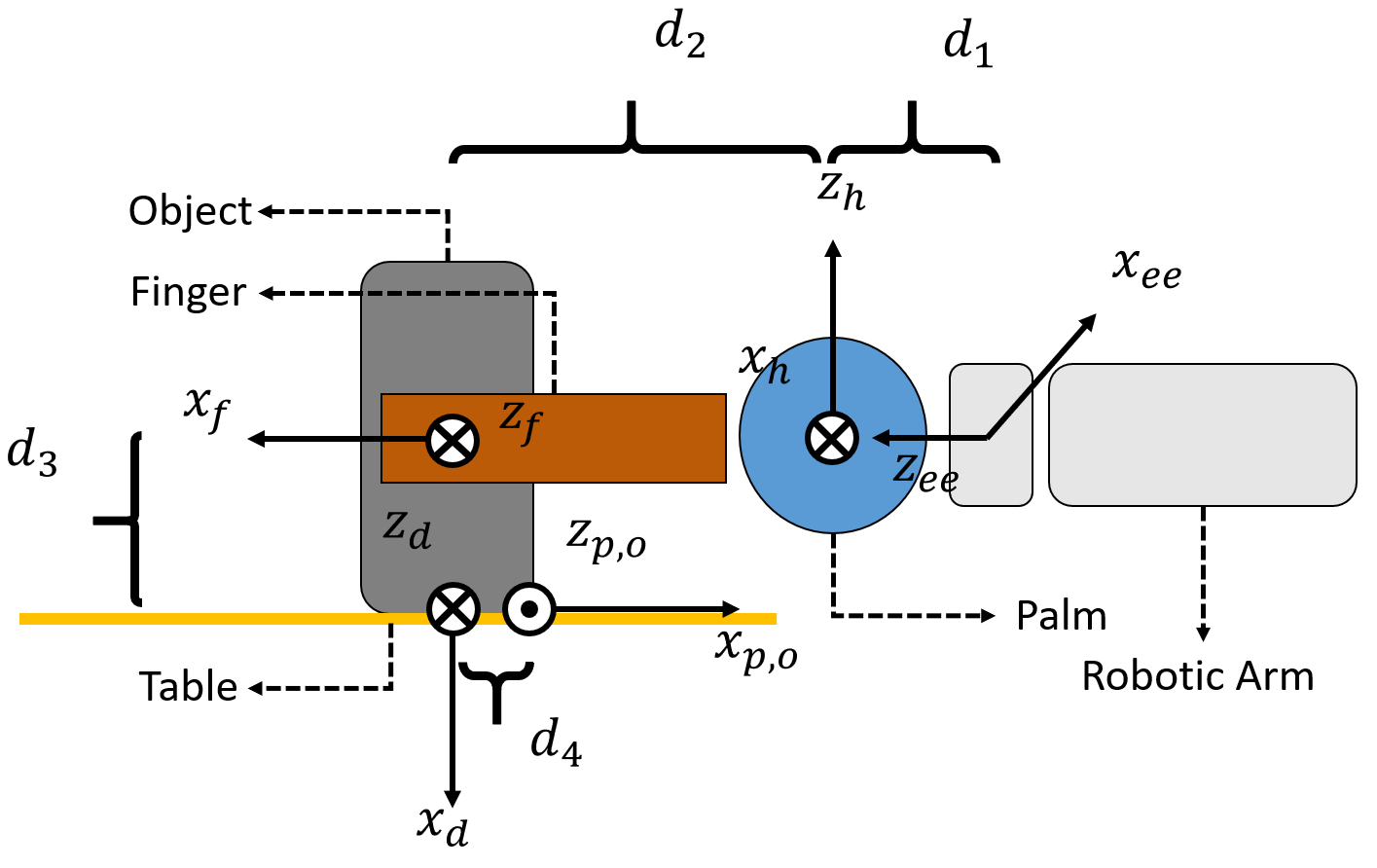}
    \vspace{-20pt}
    \caption{Frames are attached on the robot end-effector, gripper palm, finger contact, object and pivoting center. Parameter $d_1$ is a fixed offset depending on the arm-gripper setup. Planning algorithm outputs the current finger parameters ($\theta_{finger}$, $d_2$, $d_3$ and $d_4$) for a pivoting action, which are then used in the kinematic model. During the pivoting, finger parameters are held constant, while the angles of the virtual joints ($\theta_{contact}$ and $\theta_{pivot}$) vary.}
    \label{fig:pivoting_frame_assignment}
    \vspace{-5pt}
\end{figure}

\begin{figure}[t!]
    \centering
    \includegraphics[width=\linewidth]{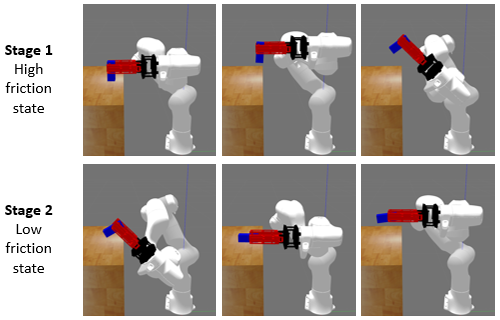}
    \vspace{-20pt}
    \caption{Pivoting is executed in two stages. At Stage 1, fingers are in high friction state, as the object rotates around the pivoting axis. At Stage 2, object is  simultaneously rotated around the pivoting axis and finger contact by utilizing the table surface and switching to low friction state.}
    \label{fig:pivoting_stages}
    \vspace{-20pt}
\end{figure}

Similarly, being able to change finger surface friction allows us to leverage extrinsic contacts reliably without the need of force control or dynamic models. A kinematic model is derived by modelling contacts between the fingers and the object and between the object and the supporting surface as revolute joints. Frame assignments are shown in Fig. \ref{fig:pivoting_frame_assignment}. Using the Denavit-Hartenberg parameters provided in Table \ref{table:pivoting_DH}, transformations between the end-effector frame and the object frame is found. According to the kinematic model, pivoting is executed in two stages as shown in Fig. \ref{fig:pivoting_stages}:
\begin{enumerate}
    \item Maintaining the contact angle with the fingers at high friction state, object is rotated around the pivoting axis by controlling the end-effector pose via robot motions.
    \item Keeping the high friction state, the robot moves the object to establish a contact with the support surface at the pivoting point. The hand switches to low friction state for both fingers, and the robot follows an arch motion based on the derived kinematics.
\end{enumerate}

\renewcommand{\arraystretch}{1.5}
\begin{table}[t!]
    \caption{DH Parameters for arm-hand system}
    \vspace{-10pt}
    \centering
    \begin{tabular}{|c|c|c|c|c|}
         \hline
         Transformation & $\theta$ & $d$ & $a$ & $\alpha$ \\
         \hline
         $T^{ee}_h$ & $\frac{3\pi}{4}$ & $d_1$ & 0 & $\frac{\pi}{2}$ \\
         \hline
         $T^{h}_f$ & $\theta_{finger}$ & 0 & $d_2$ & $\frac{\pi}{2}$ \\
         \hline
         $T^{f}_d$ & $\theta_{contact}-\frac{\pi}{2}$ & 0 & $d_3$ & 0 \\
         \hline
         $T^{d}_p$ & $-\frac{\pi}{2}$ & 0 & $d_4$ & $\pi$ \\
         \hline
         $T^{p}_o$ & $\theta_{pivot}$ & 0 & 0 & 0 \\
         \hline
    \end{tabular}
    \label{table:pivoting_DH}
     \vspace{-20pt}
\end{table}
\renewcommand{\arraystretch}{1}


\section{Region-Based Motion Planning}
\label{sec:region-based_planning}

Considering a within-hand manipulation task that requires the contact between the hand and the object to be navigated to a desired region on the object, we develop a region-based motion planning formulation.

\subsection{Motion Planning Problem Formulation}
\label{sec:motion_planning_problem}

Region-based within-hand manipulation planning problem is defined as finding a sequence of manipulation primitives $\pi=[a_1, ..., a_T]$ that navigates the contact between the hand and the object from an initial state $s_0$ (the initial contact region) to a set of desired regions $P=\{p_1,...,p_n\}$ at step $T$. Regions $p_1$ to $p_n$ are on different surfaces of the object through $1-n$. For the variable friction hand, there is a contact region between each finger and the object to be navigated within one of the goal regions. In this work, we have the following assumptions:

\begin{enumerate}
    \item 3D geometry of the object is assumed to be known. 
    \item Object surfaces and contact regions are convex polygons.
    \item Fingers make contact with two parallel surfaces on the object with opposing normals.
\end{enumerate}

Our motion planning algorithm aims to solve the optimization problem given as:
\vspace{-6pt}
\begin{equation}
    \min_{\pi} (E(s_T,P) + w \sum^{T-1}_{t=0}g(a_t))
\vspace{-6pt}
\end{equation}
where $E(s_T,P)$ is a cost function designed to minimize the contact region that is left outside of the goal region at step $T$, $g(a_t)$ is the cost of taking an action $a$ at time $t$ and $w$ is a trade-off weight between the cost components. Depending on the limitations in state space, kinematic model and hardware, constraints can be imposed on the optimization problem.

There are 6 manipulation primitives performed by the variable friction hand. 3 additional manipulation primitives are enabled by the cooperation of the robotic arm and the hand. Total of 9 actions are considered in the motion planning problem, resulting in the following action space: $A =$ \{Slide along the left finger up, slide along the left finger down, slide along the right finger up, slide along the right finger down, rotate clockwise, rotate counterclockwise, move contact up (along the z direction in Fig.~\ref{fig:DOF}b), move contact down, pivot\}.

Each of these actions are discretized as follows. The resolution of slides and moving contact up-down are dependent upon the range of dimensions of the objects being manipulated. Also, having a small resolution creates a burden for the planning algorithm and increases the planning time of the required manipulation sequence. Based on these factors, the resolution is determined experimentally. The resolution of in-hand rotation and pivoting actions are determined automatically according to the object geometry.

\begin{figure}[t!]
    \centering
    \includegraphics[width=6cm]{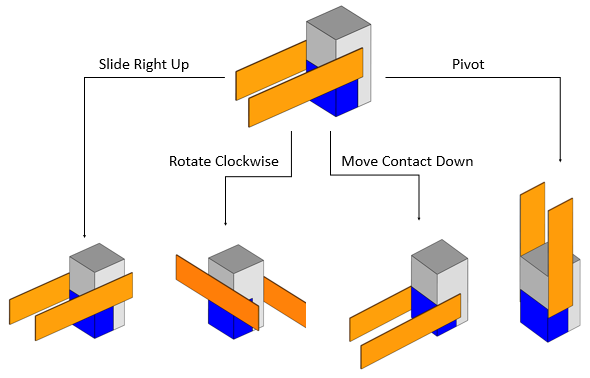}
    \vspace{-10pt}
    \caption{When an action is taken, contact regions are transformed according to a transition model. Following an action, the contact region can translate and rotate on the same surface or travel between different surfaces.}
    \label{fig:transition_model}
    \vspace{-20pt}
\end{figure}

Our transition model illustrated in Fig. \ref{fig:transition_model} describes how taking an action $a$ transforms the current state $s$ to the next state $s'$ (e.g. how pivoting the object changes the contact region between the fingers and the object). For any given object and a set of goal regions complying with our assumptions, proposed algorithm can automatically determine the valid manipulation actions and use the transition model to compute the contact states, during the search for a solution sequence.

\subsection{Cost Functions and A* Search}
\label{sec:cost_functions}

We solve the optimization problem defined in Section~\ref{sec:motion_planning_problem} with a modified A* algorithm and a region-based heuristic design.

\begin{figure}[t!]
    \centering
    \includegraphics[width=\linewidth]{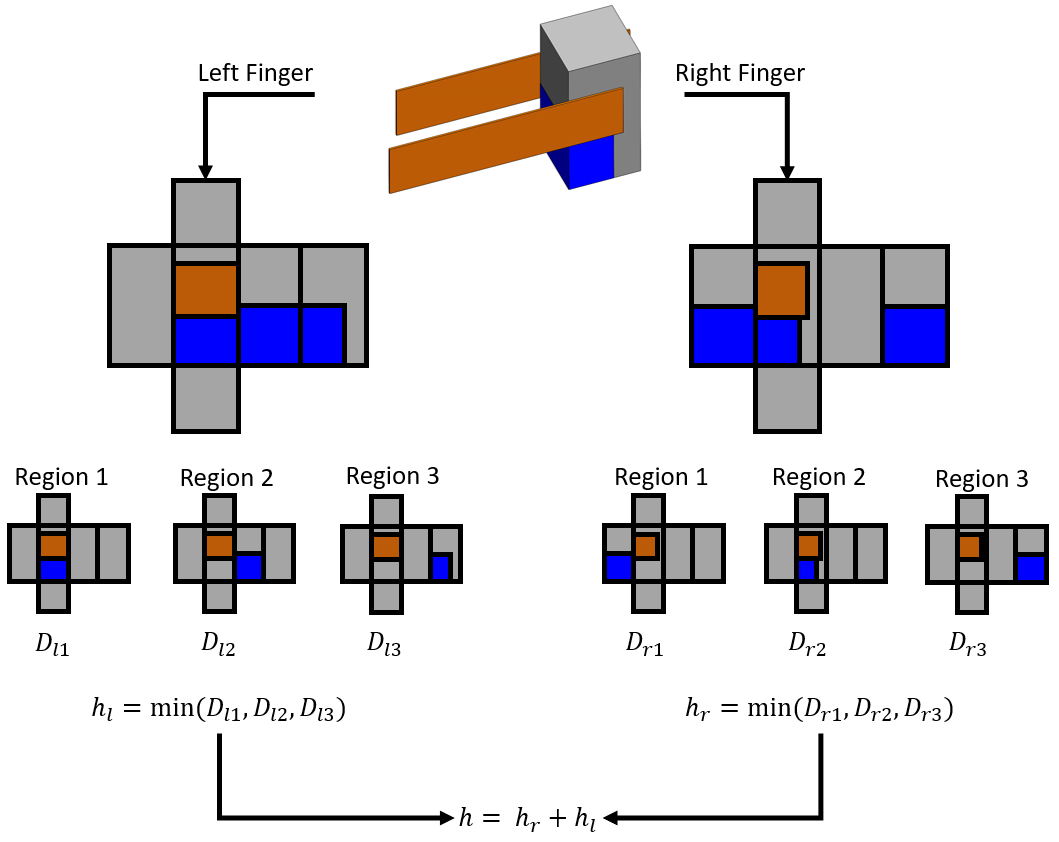}
    \vspace{-20pt}
    \caption{To compute the heuristic function, surfaces of the object are unfolded, while keeping the contact surface at the origin. For each goal region the shortest distances with the contact are computed. Minimum of the shortest distances is the heuristic for the selected finger.}
    \label{fig:heuristic_design}
    \vspace{-20pt}
\end{figure}

The heuristic function is used to attract the search for the manipulation sequence towards the goal regions. Therefore, it includes a measure of distance between the contact regions and the goal regions on the object surface. Since manipulation primitives move the contact region along the surface of the object, distances to be measured also needs to be along the object surface. Geodesic distance computation methods are developed for this purpose \cite{Crane:2017:HMD}. However, a quicker and simpler approach is required to be implemented in the proposed planning algorithm. We propose a method that automatically unfolds the surfaces of a given object and projects them onto a plane. This approach ensures that the euclidean distances measured on the projection plane are the same with the distances that are measured along the object surface. Fig. \ref{fig:heuristic_design} illustrates the unfolding procedure and the computation of the heuristic.




Let $c_i$ denote the $i$-th vertex of the contact region between the fingers and the object, and $p_m$ denotes the goal region on surface $m$. The shortest distance between a point and the surface is computed with function $d$. Following steps summarize the details of the heuristic computation:
\begin{enumerate}
    \item For each corner on a contact region ($c_i$), the shortest distance between the corner and the goal region on a selected surface is computed ($d(c_i,p_m)$). Distances are summed up for 4 corners.
    \vspace{-6pt}
    \begin{equation}
        D_m = \sum^4_{i=1} d(c_i,p_m)
    \vspace{-6pt}
    \end{equation}
    \item Step 1 is repeated for each goal region, $P=\{p_1,...,p_n\}$.
    \item Minimum of the stored sums is the heuristic at state $s$ for the selected contact region.
    \vspace{-6pt}
    \begin{equation}
        h(s) = \min(D_1,D_2,...,D_n)
    \vspace{-6pt}
    \end{equation}
    \item Heuristic is computed for both right and left contact regions. Final heuristic is the sum of the right and left heuristics.
    \vspace{-6pt}
    \begin{equation}
        h_{final}(s) = h_l(s) + h_r(s) 
    \vspace{-6pt}
    \end{equation}
\end{enumerate}


While determining the costs of manipulation primitives, we consider the in-hand sliding as the reference case, where the cost of a unit sliding is equal to the sliding resolution. Costs of the remaining primitives are scaled up according to the required arm motion and time to execute. Therefore, the distance measure used as the heuristic is designed to be an optimistic estimate of the required cost for admissibility.

\section{Experimental Results}
\label{sec:experimental_results}

\begin{figure}[t!]
    \centering
    \includegraphics[width=6cm]{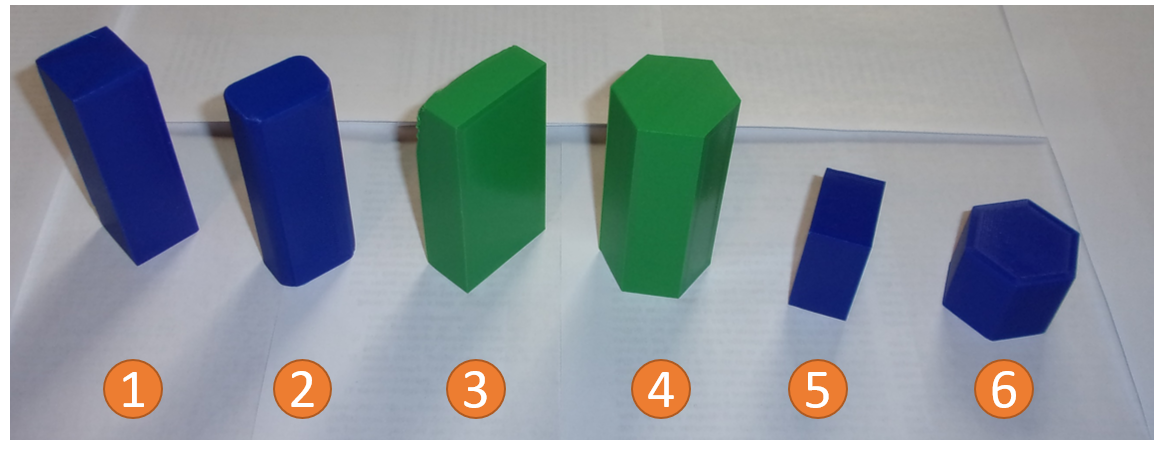}
    \vspace{-10pt}
    \caption{Artificial objects are modeled and 3D-printed for the experiments. Set of objects include: 1) square prism 2) rectangular prism curved 3) rectangular prism large 4) hexagonal prism tall 5) rectangular prism small 6) hexagonal prism short.}
    \label{fig:object_set}
\end{figure}

\begin{figure}[t!]
    \centering
    \includegraphics[width=\linewidth]{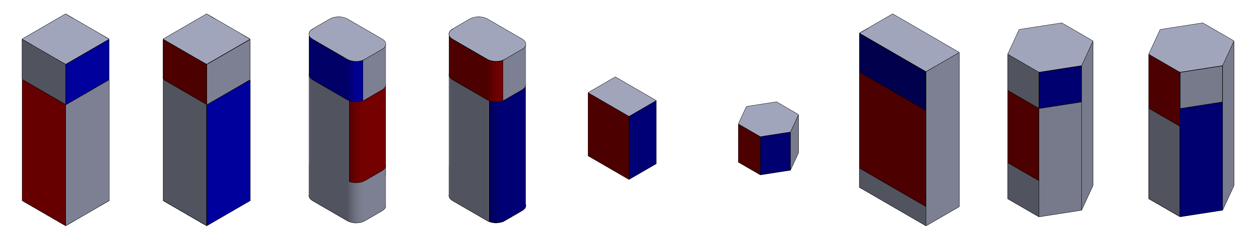}
    \vspace{-20pt}
    \caption{Initial and goal region pairs are defined on each object. Red regions indicate the initial regions and blue regions indicate the goal regions.}
    \label{fig:goal_regions}
    \vspace{-10pt}
\end{figure}

To validate our approach and evaluate the manipulation sequences generated by the planning algorithm, we conduct real-robot experiments using the variable friction hand setup and a Franka Emika Panda arm. To be used in the experiments, we designed and manufactured the object set given in Fig. \ref{fig:object_set} from polylactic acid (PLA) plastic by 3D printing\footnote{The main logic of the variable friction finger design is to have the friction coefficients of the high and low friction surfaces at extremes, which was sufficiently achieved by the materials used in the current fingers. This property enables controlled sliding and rotation of a wide range of objects with different friction properties.}. For each object, 1-3 initial and goal regions are defined. Some examples are shown in Fig. \ref{fig:goal_regions}

We use the in-hand manipulation benchmarking protocol proposed in \cite{cruciani-ral2020-benchmarking-in-hand}. 1-3 manipulation sequences are generated for each object corresponding to each initial and goal region. 5 trials are run for each generated plan, with a total of 5-15 executions per object. Average motion planning and execution times are provided along with the failure rates for each object in Table \ref{table:results}.


\renewcommand{\arraystretch}{1.5}
\begin{table}[t!]
    \caption{Experiment Metrics}
    \vspace{-10pt}
    \centering
    \begin{tabular}{cp{1cm}p{1cm}p{1cm}}
         \toprule
         Object & Planning time (s) & Execution time (s) & Failure rate \\
         \midrule
         \textit{square prism} & 70.8 & 70.7 &26\% \\
         \hline
         \textit{rectangular prism curved} &  40.1 & 52.3 & 10\%\\ 
         \hline
         \textit{rectangular prism large} & 34.1 &  39.5 & 10\%\\
         \hline
         \textit{hexagonal prism tall} & 5.7 & 17.1 & 0\% \\
         \hline
         \textit{hexagonal prism short} & 8.5 & 10.0 & 0\% \\
         \hline
         \textit{rectangular prism small} & 1.4 & 10.9 & 0\% \\
         \hline
         \textbf{\textit{All Objects}} & 37.6 & 40.8 & 10\%\\
         \bottomrule
    \end{tabular}
    \label{table:results}
    \vspace{-15pt}
\end{table}
\renewcommand{\arraystretch}{1.5}

\begin{figure}[t!]
    \centering
    \includegraphics[width=\linewidth]{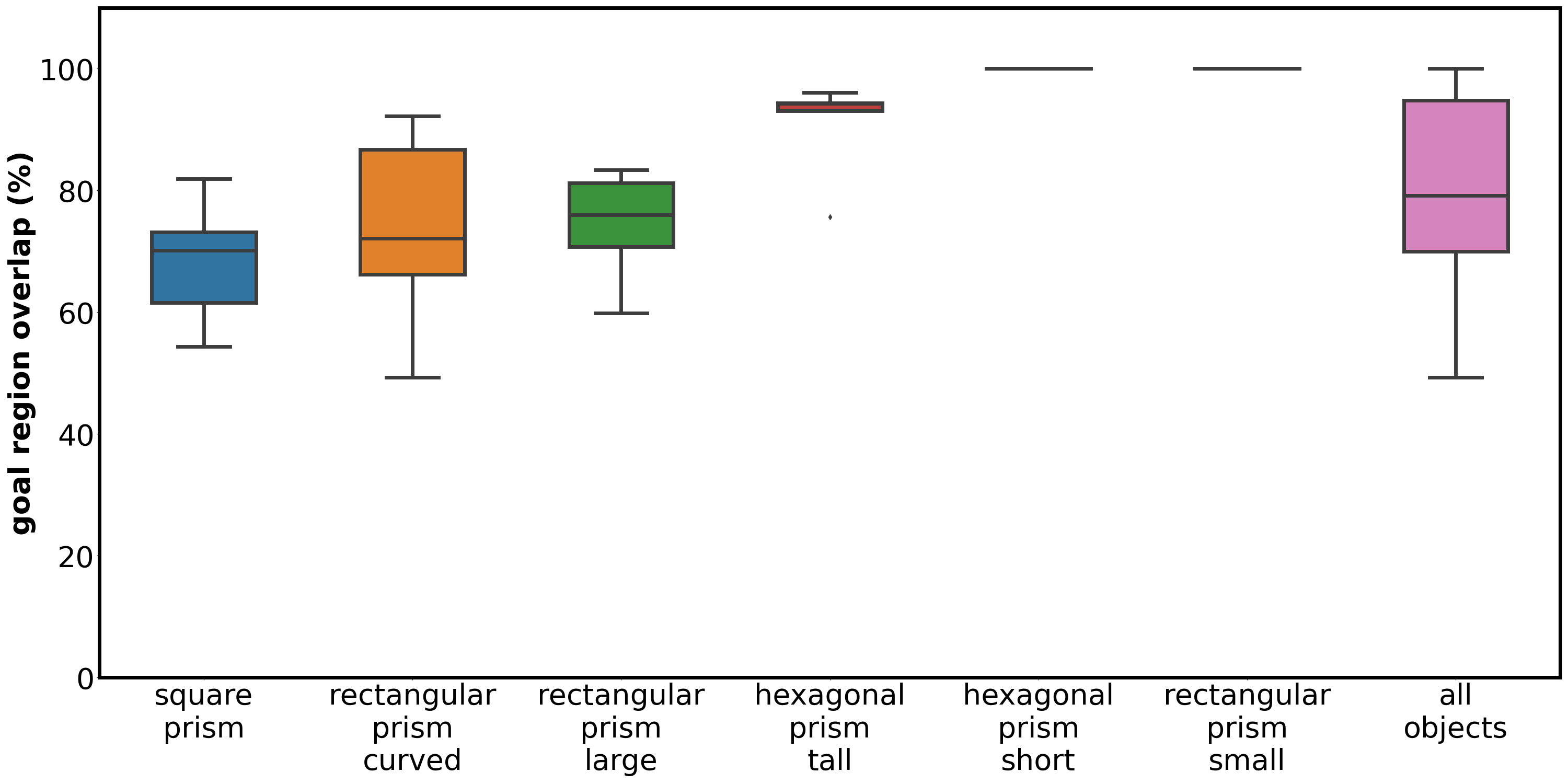}
    \caption{Distribution of averaged goal region overlap for left and right fingers on each object. Results include 11 successful trials for square prism, 9 successful trials for rectangular prism curved and rectangular prism large, 5 successful trials for hexagonal prism tall, hexagonal prism short and rectangular prism small for a total of 44 successful trials.}
    \label{fig:results}
    \vspace{-20pt}
\end{figure}

To evaluate the performance of our system in region-based manipulation tasks, we compute the ratio of the contact area within the goal to the area of the entire contact region. Computed percentage goal region overlap is reported in Fig. \ref{fig:results}.

Results show that the region-based planning and the within-hand manipulation pipeline can navigate around 70\% of the contact region into the goal region for the selected objects. Errors in the final state are mostly due to the open-loop execution of the manipulation primitives, which fails to compensate for the inaccuracies in the modeling and disturbances induced by the unplanned slips during the execution. In some cases, unplanned slips lead to dropping the object, which are referred to as failures in Table~\ref{table:results}. The performance and the failure rates are highly dependent upon the complexity of the defined regions on the object and corresponding manipulation plans. Square prism, being the object with the most primitive geometry, allows more complex manipulation planning and execution including multiple slides and rotations in sequence with pivoting. However, likelihood of having unplanned slippage during manipulation also increases with the number and complexity of the executed manipulation actions, which caused larger region overlap errors. Curved and large rectangular prisms were subject to less complex sequences including slides with single rotation and pivoting towards the end. Hexagonal prisms and small rectangular prism were not pivoted during the experiments due to the workspace constraints, which reduced the corresponding failure rates and inaccuracies drastically for these objects. The effect of manipulation complexity can also be observed on the planning and execution times provided in Table~\ref{table:results}.

\section{Discussion and Future Work}
\label{sec:discussion}

While the planning strategy, coupled with the advantages of the variable friction fingers, provides a large operation space, the setup exhibits several mechanical limitations that prevent the execution of manipulation primitives on certain objects or certain states. These limitations sometimes cause inaccuracies or failures during the execution. We accounted for some of these limitations by imposing constraints on the optimization problem (e.g. by introducing some action sequences as workspace constraints, making the system avoid near-failure states). These constraints are parametrically defined inequalities depending on the object and finger geometry. However, the hand was not able to manipulate some object geometries at all, for instance, thin prisms and moderately large objects, which we removed from our experiment set. According to our observations, the major limiting factor was the palm width of the hand. For objects with certain length to width ratios, it is not always possible to exert enough moment to initiate the within-hand rotation motion. Similarly, for some object geometries and palm width, a slight disturbance or modeling mismatch can initiate a rotation, preventing the sliding actions to be executed as planned. In the future work, we plan to investigate the advantages of having a variable palm width that can be actively actuated during the manipulation. In addition to enabling sliding and rotation for thin prisms, this would also enable the manipulation of larger or smaller objects. However, it is expected that such modifications will increase the system complexity and motion planning requirements.

Another potential direction with this work is to incorporate closed-loop control and planning to the proposed within-hand manipulation framework. By estimating the object pose within-hand, we plan to account for the modeling errors in variable friction hand kinematics as well as the pivoting kinematics. This might enable us to avoid near failure configurations and reach goal regions with increased accuracy.

\section{Conclusion}
\label{sec:conclusion}

Based on the variable friction fingers introduced in prior work, we propose an automatic 3D within-hand manipulation platform that consists of an arm-gripper system and a region-based planning algorithm. Using the variable friction principle, we proposed a quasi-static approach to extrinsic dexterity based manipulation strategies such as prehensile pushing, gravity exploitation and pivoting. Relying on the contact mode control achieved through friction modulation at the finger surfaces, manipulation primitives are modeled in purely kinematic-level and executed without elaborate sensing or control. We consider a region-based within-hand manipulation task and formulate it as an optimization problem. Developed  region-based motion planning algorithm uses modified A* search to solve the motion planning problem and generate a sequence of manipulation primitives to move the fingers from an initial contact region to a given goal region. Our approach is validated through real robot experiments following a standardized in-hand manipulation benchmarking protocol.


\addtolength{\textheight}{-13.5cm}  
\bibliographystyle{IEEEtran}


\end{document}